\documentclass[sigchi]{acmart}
\renewcommand\footnotetextcopyrightpermission[1]{} 
\AtBeginDocument{%
  \providecommand\BibTeX{{%
    \normalfont B\kern-0.5em{\scshape i\kern-0.25em b}\kern-0.8em\TeX}}}

\usepackage{pgfplots}
\pgfplotsset{compat=newest}
\usepgfplotslibrary{groupplots} 
\usepgfplotslibrary{dateplot}
\usepackage{mdframed}
\usepackage{CJKutf8}

\begin{document}

\setcopyright{none}
\settopmatter{printacmref=false}
\renewcommand\footnotetextcopyrightpermission[1]{}
\pagestyle{plain}

\title[Transportation Mode Classification from Smartphone Sensors…]{Transportation Mode Classification from Smartphone Sensors via a Long-Short-Term-Memory Network}



\author{Bj\"{o}rn Friedrich}
\affiliation{%
  \institution{Carl von Ossietzky University}
  \city{Oldenburg}
  \country{Germany}
  }
\authornote{Both authors contributed equally to this research.}

\author{Benjamin Cauchi}
\authornotemark[1]
\affiliation{%
  \institution{OFFIS e.V.\\ Institute for Information Technology}
  \city{Oldenburg}
  \country{Germany}
  }
  
\author{Andreas Hein}
\affiliation{%
  \institution{Carl von Ossietzky University}
  \city{Oldenburg}
  \country{Germany}
 }
  
\author{Sebastian Fudickar}
\affiliation{%
  \institution{Carl von Ossietzky University}
  \city{Oldenburg}
  \country{Germany}
  }

\renewcommand{\shortauthors}{Friedrich and Cauchi et al.}

\begin{abstract}
  This article introduce the architecture of a Long-Short-Term-Memory network for classifying transportation-modes via smartphone data and evaluates its accuracy. By using a Long-Short-Term-Memory with common preprocessing steps such as normalisation for classification tasks an F1-Score accuracy of 63.68\ \%  was achieved with an internal test dataset.\\
  We participated as team "\emph{GanbareAMT}" in the “\emph{SHL recognition challenge}".
\end{abstract}





\maketitle
\section{Introduction}
Knowledge about the specific modes of transportation in urban areas are very important. More and more people are moving to urban areas and stress traffic infrastructure and especially the means of mass transportation. Knowledge about the mode of transportation and the flow of people inside a traffic system is important for optimising the traffic system, traffic control, travel behaviour research and the recommendation of travelling routes and means of transportation for people themselves.\\
Such information is typically collected via questionnaires and phone-surveys. However, these methods are time consuming and expensive - since requiring manual labor. Since smartphones are omnipresent it is reasonable to use the data collected by smartphone-sensors to identify mode of transportation.

Thus, the recognition of transportation modes is a relevant step to enhance the contextual knowledge of user, as required for behavioural investigations and characterisation of the users physically activity level, among others. 
With over 700 hours of annotated data covering 8 transportation modes, which have been recorded by 3 participants, the "\emph{Sussex-Huawei Locomotion- Transportation (SHL) dataset}" \cite{RoggenDatasetAndSurvey, SHLDataset1} represents a milestone for the development and training of algorithms for locomotion analysis via mobile devices. 
Due to its coverage of multiple sensor-types (such as inertial sensors, magnetometer and GPS) and wearing positions it is well suited to compare the accuracy among algorithms transportation mode classifiers and sensor-modalities. 

By setting up the Sussex-Huawei locomotion machine learning/data science challenge 2019, Roggen and colleagues challenge define a venue to compare the efficiency of various approaches to classify 8 modes of locomotion and transportation via inertial data from smartphone-integrated sensory. Thereby, the variation of sensor placement on various body-parts among training-set and test-set is foreseen as the main challenge, Since previously studies have indicated that the classification of movements via inertial data is mainly dependent on sensor-placement \cite{KunzeSensorVariations}.

The article at hand introduces one training network, that has been proposed for the challenge we participated in as team "\emph{GanbareAMT}".

\section{Related Work}
Large scale datasets for working with mode of transportation tasks are rare. One of the largest ones is the Microsoft Asia Geolife Trajectory dataset. The data has been collected over four years with 178 users \cite{zheng2011geolife}. With a variety of sensor models being used, the dataset only contains GPS data and therefore it is not comparable to the dataset for the challenge. 
As pointed out in \cite{RoggenDatasetAndSurvey} various sensor-types, features, signal-preprocessing methodologies and  machine learning (ML) methods have been shown to be suitable for transportation mode and human activity recognition. The comprehensive summary shows that besides the classic  algorithms like Support-Vector-Machines and k-Nearest-Neighbours deep learning techniques have been used as well. Even though the suitabilty to classify human activity recognition on IMU data via  Recurrent Neural Networks and Long-Short-Term-Memory (LSTM) networks are confirmed \cite{lstmApproach}, they have not been considered in the review.\\
E.g. Zebin et al. acquired 92 \% accuracy by classifying eight different human activitiesvia LSTM for IMU data \cite{lstmApproach}. Since, the accuracy of such classifyer networks be most meaningful evaluated on a common dataset, the article at hand aims to evaluate the suitability of LSTM-based network on the Sussex-Huawei Locomotion-Transportation (SHL) dataset.  

\section{Dataset}
The "Sussex-Huawei Locomotion-Transportation (SHL) data\-set"\cite{SHLDataset1,RoggenDatasetAndSurvey} features a great amount labeled data. The data has been collected by 3 participants each carrying in parallel HUAWEI Mate 9 smartphones at 4 different body-positions (hips, bag, torso and hand). The resulting 2800 hours of annotated data. The data of the accelerometer, gyroscope, magnetometer and an ambient pressure sensor included in the smartphone, as well as the orientation in quaternions ,the gravity and the linear acceleration were provided for this challenge. The IMU sensors' sampling rates is around 100\ \textit{Hz}.

The challenge-dataset represents a selected subset of the SHL dataset. Herein, the data was divided into 5 second blocks (each consequently covering 500 data points). For the training and development set, the activities within the 500 labels might alter per sample-block, while in the test-set each block seams to contain labels for only one activity.

Among the datasets made available in the session, the following speciality of the included sensor position, is relevant - representing the fact, that the challenge wants to evaluate the transferability of trained samples for other conditions. The challenge dataset features the accelerometer, gyroscope, magnetometer, orientation, gravity, linear acceleration and ambient pressure. The training set contains data of the smartphones located at the bag, torso and hips. The validation set contains not only data of the bag, torso and hip, but also from the phone carried in the hand. In contrast, the test set comprises data of the hand only - the intended evaluation position.

While the general dataset holds various labels, for the given challenge only eight labels considering the transportation mode (namely "Still", "Walking", "Run", "Bike", "Car", "Bus", "Train" and "Subway") are relevant, and thus, considered.

\begin{figure}[h]
  \centering
  \includegraphics[width=\linewidth]{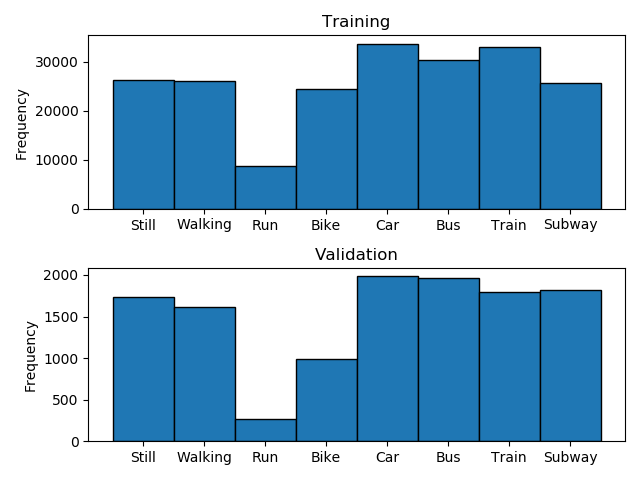}
  \caption{This figure shows the distribution of the labels in the dataset. The validation set contains data of the hand only.}
  \Description{Histogram of Labels}
  \label{fig:labelhist}
\end{figure}

The label-distribution in the given dataset (considering training and validation dataset)is imbalanced as shown in Figure \ref{fig:labelhist}. Herein sample-blocks in which labels varied, the label for the sample-block was determined per majority decision - an approach which appears appropriate since for the considered dataset - in this rare condition of varying activities, typically over 90\% belonged to one label type.

\section{Data Processing Pipeline}

Since the challenge aimed at training on the phones placed at bag, torso and hip, while  predicting on the data of the phone in the hand we combined the corresponding data in the training and validation set. Therefore, our training set comprises of the data of the 3 phones and our validation of the data of the phone carried in the hand as summarized in Table \ref{Tab:SampleDist}. 

In order to overcome the dataset's imbalance, the dataset was balanced among the classes by under-sampling (that is by excluding samples for over-represented classes until a balance is achieved). 
The combined use of under-sampling and oversampling (via duplication of "Ru" class samples) for the training- and validation-set was as well considered but did not show convincing results. 

\begin{table*}
  \caption{Sample distribution for the sensor-placement among training, development and test-data set: An equal distribution among the considered classes of transportation-mode has been assured via under-sampling.}
  \begin{tabular}{l l  c  c  c  c}
    \toprule
  	Type & Total & \multicolumn{4}{c}{Sample distribution among placement}    \\ 
	& samples & Hip & Bag & Torso & Hand  \\
	\midrule
	Training & 209280 & 69760 (33\ \%) & 69760 (33\ \%) & 69760 (33\ \%) & - \\ 
	Validation & 2136 & - & - & - & 2136 (100\ \%) \\
	Test & 55811 & - & - & - & 55811 (100\ \%) \\ 
  \bottomrule
\label{Tab:SampleDist}
\end{tabular}
\end{table*}

We are using all raw-data features provided in the dataset except for gravity and orientation, resulting in acceleration $(x,y,z)$, gyroscope $(x,y,z)$, magnetometer $(x,y,z)$, "linear acceleration" and pressure.
We are calculating the activation as a single input feature over sensors as the sum of the Euclidean distance over all three axis per sensor by equation
$$ 
\sum_{i=0}^{4}\sqrt{x_i^2+y_i^2+z_i^2} + p 
$$
where i denotes the sensor. Thereby, each datum in a sequence is composed of all sensor data available for that specific measurement time.
\begin{figure}[h]
  \centering
  \includegraphics[width=\linewidth]{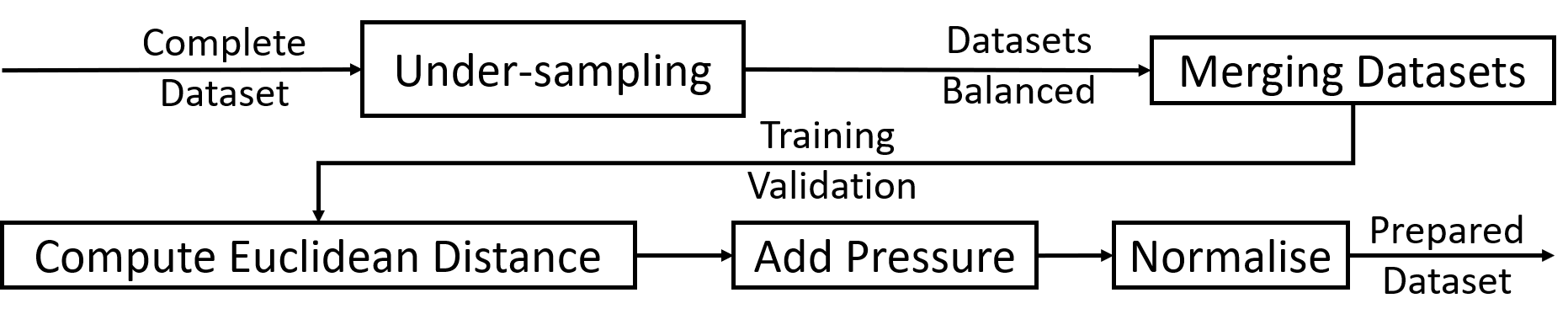}
  \caption{Our preprocessing pipeline is composed of 5 steps. The superficial samples have been removed and the data was merged from the different positions. Afterwards,the Euclidean distance is computed per sensor and pressure is added. Finally, the summarised datum is normalised.}
  \Description{Processing Pipeline}
  \label{fig:processing}
\end{figure}
The data in the validation and test set are not consecutive in time. Hence, the window size is limited to 5 seconds. We are using a window size of 1 second, so that one sample is of shape $5x100$. In terms of a LSTM input we have 100 features and 5 time steps. Before fed to the network, the dataset is normalised within a range of $(-1,1)$. Figure \ref{fig:processing} depicts the resulting 5 steps of our preprocessing pipeline.

\section{Network Architecture}
\begin{figure}[h]
  \centering
  \includegraphics[width=\linewidth]{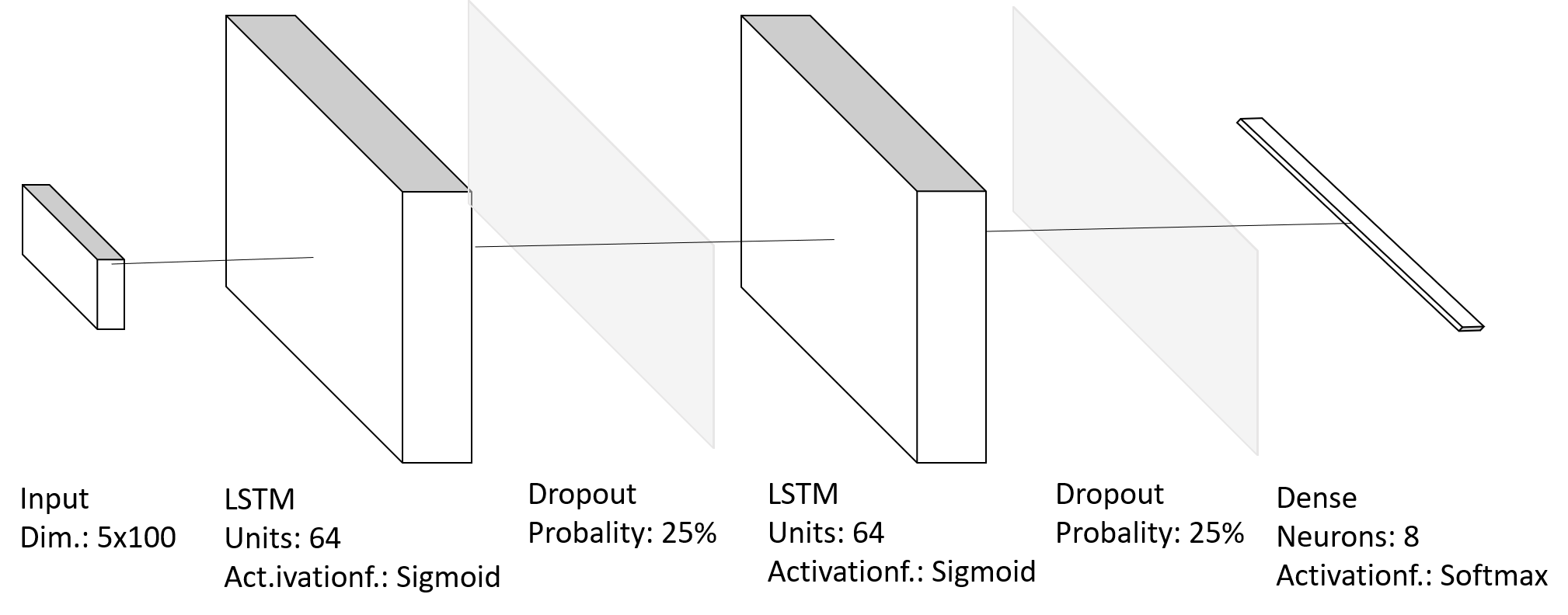}
  \caption{This figure shows the architecture of our two layer LSTM network. Due to the basic the principle of LSTMs even this small network has 51720 trainable parameters.}
  \Description{LSTM Architecture}
  \label{fig:architecture}
\end{figure}
A LSTM network has been used, due to its shown suitability for temporal data. LSTMs not prone to the vanishing gradient or exploding gradient problem compared to common recurrent neural networks \cite{LSTM, LSTMForget}.\\
Our architecture is composed of two LSTM layers and two dropout layers. On top a classification layer with eight neurons for the eight classes was added. Both LSTM layers consist of 64 neurons each and are activated by a Sigmoid activation function. The dropout layers have a probability of 25 \% for dropping a neuron. The classification layer uses the Softmax function. The resulting network has overall 51720 trainable parameters and its architecture is shown in Figure \ref{fig:architecture}.

\section{Training}
The resulting LSTM was trained for 197 epochs with the checkpoint saving condition to best, so that a checkpoint is saved, if there is an improvement in the validation accuracy. Finally, the most accurate configuration among all epochs is used. The ADAM optimizer \cite{Kingma2015AdamAM} is used with the categorical-crossentropy loss function and the corresponding Categorical Accuracy metric, due to the classification task with eight classes. We performed the experiments a NVIDIA TITAN V GPU with 12GB of VRAM as external GPU which was connected via a 40\ \emph{GBit/s} Thunderbolt3 connection - with each training epoch takes approximately 32\ \emph{minutes}. The final training was conducted on a NVIDIA Tesla P-100 with 16GB of VRAM were each training epoch takes approximately 40\ \emph{minutes}. The graphics card is in a node of a high-performance computing facility. We reserved 120GB of RAM and one Intel Xeon CPU with 12 cores at 2.2GHz as well.

\section{Results}
\begin{figure}[h]
  \centering
  \includegraphics[width=\linewidth]{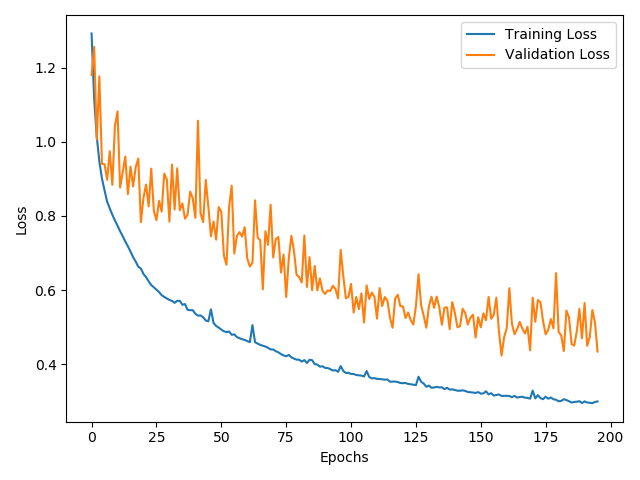}
  \caption{This figure shows the progress of the loss over training time. We see an asymptotic behaviour after epoch 175.}
  \Description{Loss over training time}
  \label{fig:lstmloss}
\end{figure}
Figure \ref{fig:lstmloss} shows the progress of the loss over training time or epochs. The training loss shows a steep decrease over the first 125 epochs. After epoch 125 the loss is still decreasing, but much slower. The training loss is steadily decreasing except for a few epochs where we can see some small outliers. The loss on the validation set is steadily decreasing as well, but shows much more fluctuations.
\begin{figure}[h]
  \centering
  \includegraphics[width=\linewidth]{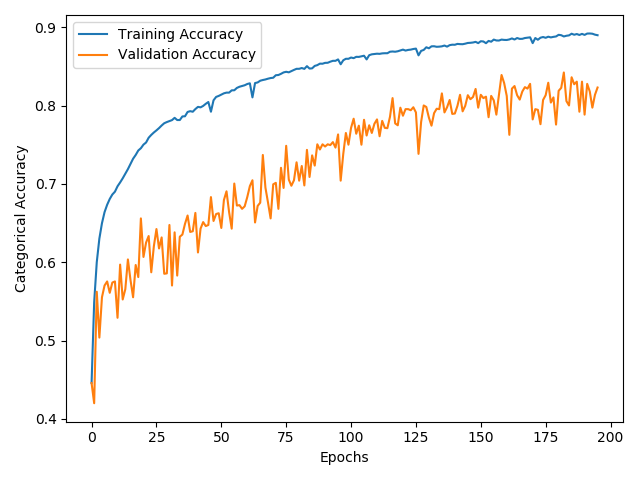}
  \caption{This figure shows the progress of the Categorical Accuracy over training time. In compliance with the loss there is an asymptotic behaviour after 175 as well.}
  \Description{Categorical Accuracy over training time}
  \label{fig:lstmacc}
\end{figure}
The progress of the accuracy goes with the progress of the loss. We see some small outliers at the same epochs. The training accuracy is steadily increasing, where the validation accuracy shows fluctuations while increasing over all.\\
For our prediction on the test set we used the epoch 183 with the best validation accuracy. At this epoch the training and validation loss were 0.3058 and 0.4357 and the accuracy 88.82\ \% for training and 84.22\ \% for validation. The predictions on the test set took about 143 \emph{seconds} on the high-performance computing cluster. The 197 epochs for training took about 4 \emph{days} 4 \emph{hours} 26 \emph{minutes} 30 \emph{seconds} on the cluster. The classification time of the test set was about 143 \emph{seconds}.
\begin{figure}[h]
  \centering
  \includegraphics[width=\linewidth]{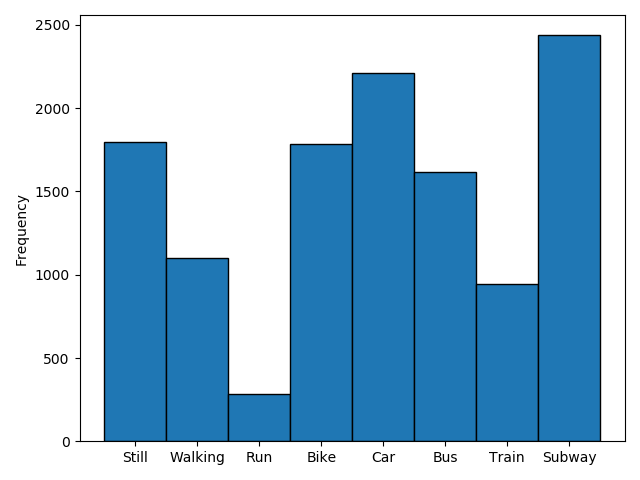}
  \caption{This figure shows the distribution of the labels on the classified test dataset. We can see some similarities between the class "Run" of the distributions in the training and validation set. It seems there are too many samples in the "Subway" class.}
  \Description{Histogram of the Classified Test Dataset.}
  \label{fig:labeldisttest}
\end{figure}
Figure \ref{fig:labeldisttest} shows the distribution of the labels on the classified test set. There are only a few similarities to the distribution in the training and validation set. We can see, that the class "Run" seems to be under-represented as in the other sets and the difference between "Bike" and "Bus" is nearly the same as in the training set. The class "Subway" seems to be over-represented.\\

The confusion matrix (table \ref{Tab:confMatrix}) is computed on the unbalanced hand data of the validation set. Most false classification occurred for "Train" being classified as "Subway". On the one hand this seems reasonable, because the movement characteristics of these activity-pairs are similar and thus challenging the differentiation of the activities based on the considered input features. On the other hand, the amount of samples of "Subway" classified as "Train" is very small. Moreover, we see a few false classifications as "Still" in all classes except for "Bike". This might happen, because all other means of transportation are frequently pausing for some time (e.g. on traffic lights or subway stations) and might switch over to a Still movement pattern during these times.

\begin{table*}
  \caption{Confusion matrix summarising the performance on the unbalanced hand data of the validation set.}
  \begin{tabular}{c c c c c c c c c }
  \toprule
     & Still & Walking & Run & Bike & Car & Bus & Train & Subway \\ 
    Still & 1339 & 12 & 0 & 171 & 56 & 33 & 45 & 81 \\ 
    Walking & 15 & 1057 & 16 & 477 & 19 & 14 & 3 &8 \\ 
    Run & 0 & 2 & 262 & 2 & 0 & 1 & 0 & 0 \\ 
    Bike & 61 & 6 & 1 & 683 & 205 & 32 & 1 & 4 \\
    Car & 93 & 2 & 1 & 209 & 1292 & 289 & 38 & 62 \\ 
    Bus & 86 & 16 & 2 & 176 & 420 & 1168 & 62 & 37 \\ 
    Train & 100 & 2 & 0 & 31 & 65 & 48 & 628 & 920\\
    Subway & 105 & 4 & 0 & 38 & 153 & 31 & 167 & 1326 \\
    \bottomrule
\label{Tab:confMatrix}
\end{tabular}
\end{table*}

\section{Conclusions}
The article introduces the submission of the team "\emph{GanbareAMT}" for the SHL challenge 2019. In order to classify the mode of transportation (differentiating eight transportation modes) using smartphone-sensors, an LSTM network was trained via data-samples recorded on the hip, in a backpack and at the torso. Regarding the sensitivity, the network achieved an F1-Score accuracy of 63.68\ \%  on the hand data of the validation set. The recognition result for the testing dataset will be presented in the summary paper of the challenge \cite{SHLSummary}.

\begin{acks}
This work has been conducted as part of the projects Audio-PSS and Safety4Bikes funded by the\\ \emph{Federal Ministry of Education and Research} under grant number 02K16C201 and 16SV7668.\\
Moreover, we gratefully acknowledge the support of NVIDIA Corporation with the donation of the TITAN V GPU used for this research.
\end{acks}

\bibliographystyle{utilities/ACM-Reference-Format}
\bibliography{utilities/sample-base}


\begin{thebibliography}{9}


\ifx \showCODEN    \undefined \def \showCODEN     #1{\unskip}     \fi
\ifx \showDOI      \undefined \def \showDOI       #1{#1}\fi
\ifx \showISBNx    \undefined \def \showISBNx     #1{\unskip}     \fi
\ifx \showISBNxiii \undefined \def \showISBNxiii  #1{\unskip}     \fi
\ifx \showISSN     \undefined \def \showISSN      #1{\unskip}     \fi
\ifx \showLCCN     \undefined \def \showLCCN      #1{\unskip}     \fi
\ifx \shownote     \undefined \def \shownote      #1{#1}          \fi
\ifx \showarticletitle \undefined \def \showarticletitle #1{#1}   \fi
\ifx \showURL      \undefined \def \showURL       {\relax}        \fi
\providecommand\bibfield[2]{#2}
\providecommand\bibinfo[2]{#2}
\providecommand\natexlab[1]{#1}
\providecommand\showeprint[2][]{arXiv:#2}

\bibitem[\protect\citeauthoryear{{Gers}, {Schmidhuber}, and {Cummins}}{{Gers}
  et~al\mbox{.}}{1999}]%
        {LSTMForget}
\bibfield{author}{\bibinfo{person}{F.~A. {Gers}}, \bibinfo{person}{J.
  {Schmidhuber}}, {and} \bibinfo{person}{F. {Cummins}}.}
  \bibinfo{year}{1999}\natexlab{}.
\newblock \showarticletitle{Learning to forget: continual prediction with
  LSTM}. In \bibinfo{booktitle}{\emph{1999 Ninth International Conference on
  Artificial Neural Networks ICANN 99. (Conf. Publ. No. 470)}},
  Vol.~\bibinfo{volume}{2}. \bibinfo{pages}{850--855 vol.2}.
\newblock
\showISSN{0537-9989}
\urldef\tempurl%
\url{https://doi.org/10.1049/cp:19991218}
\showDOI{\tempurl}


\bibitem[\protect\citeauthoryear{Gjoreski, Ciliberto, Wang, Morales, Mekki,
  Valentin, and Roggen}{Gjoreski et~al\mbox{.}}{2018}]%
        {SHLDataset1}
\bibfield{author}{\bibinfo{person}{Hristijan Gjoreski},
  \bibinfo{person}{Mathias Ciliberto}, \bibinfo{person}{Lin Wang},
  \bibinfo{person}{Francisco Javier~Ordonez Morales}, \bibinfo{person}{Sami
  Mekki}, \bibinfo{person}{Stefan Valentin}, {and} \bibinfo{person}{Daniel
  Roggen}.} \bibinfo{year}{2018}\natexlab{}.
\newblock \showarticletitle{The university of sussex-huawei locomotion and
  transportation dataset for multimodal analytics with mobile devices,}.
\newblock \bibinfo{journal}{\emph{IEEE Access}}  \bibinfo{volume}{6}
  (\bibinfo{date}{23 July} \bibinfo{year}{2018}),
  \bibinfo{pages}{42592--42604}.
\newblock


\bibitem[\protect\citeauthoryear{Hochreiter and Schmidhuber}{Hochreiter and
  Schmidhuber}{1997}]%
        {LSTM}
\bibfield{author}{\bibinfo{person}{Sepp Hochreiter} {and}
  \bibinfo{person}{Jürgen Schmidhuber}.} \bibinfo{year}{1997}\natexlab{}.
\newblock \showarticletitle{Long Short-Term Memory}.
\newblock \bibinfo{journal}{\emph{Neural Computation}} \bibinfo{volume}{9},
  \bibinfo{number}{8} (\bibinfo{year}{1997}), \bibinfo{pages}{1735--1780}.
\newblock
\urldef\tempurl%
\url{https://doi.org/10.1162/neco.1997.9.8.1735}
\showDOI{\tempurl}
\showeprint{https://doi.org/10.1162/neco.1997.9.8.1735}


\bibitem[\protect\citeauthoryear{Kingma and Ba}{Kingma and Ba}{2015}]%
        {Kingma2015AdamAM}
\bibfield{author}{\bibinfo{person}{Diederik~P. Kingma} {and}
  \bibinfo{person}{Jimmy Ba}.} \bibinfo{year}{2015}\natexlab{}.
\newblock \showarticletitle{Adam: A Method for Stochastic Optimization}.
\newblock \bibinfo{journal}{\emph{CoRR}}  \bibinfo{volume}{abs/1412.6980}
  (\bibinfo{year}{2015}).
\newblock


\bibitem[\protect\citeauthoryear{{Kunze} and {Lukowicz}}{{Kunze} and
  {Lukowicz}}{2014}]%
        {KunzeSensorVariations}
\bibfield{author}{\bibinfo{person}{K. {Kunze}} {and} \bibinfo{person}{P.
  {Lukowicz}}.} \bibinfo{year}{2014}\natexlab{}.
\newblock \showarticletitle{Sensor Placement Variations in Wearable Activity
  Recognition}.
\newblock \bibinfo{journal}{\emph{IEEE Pervasive Computing}}
  \bibinfo{volume}{13}, \bibinfo{number}{4} (\bibinfo{date}{Oct}
  \bibinfo{year}{2014}), \bibinfo{pages}{32--41}.
\newblock
\showISSN{1536-1268}
\urldef\tempurl%
\url{https://doi.org/10.1109/MPRV.2014.73}
\showDOI{\tempurl}


\bibitem[\protect\citeauthoryear{{Wang}, {Gjoreski}, {Ciliberto}, {Lago},
  {Murao}, {Okita}, and {Roggen}}{{Wang} et~al\mbox{.}}{2019a}]%
        {SHLSummary}
\bibfield{author}{\bibinfo{person}{L. {Wang}}, \bibinfo{person}{H. {Gjoreski}},
  \bibinfo{person}{M. {Ciliberto}}, \bibinfo{person}{P. {Lago}},
  \bibinfo{person}{K. {Murao}}, \bibinfo{person}{T. {Okita}}, {and}
  \bibinfo{person}{D. {Roggen}}.} \bibinfo{year}{2019}\natexlab{a}.
\newblock \showarticletitle{Summary of the Sussex-Huawei
  locomotion-transportation recognition challenge 2019}.
\newblock \bibinfo{journal}{\emph{Proc. HASCA 2019}}.
\newblock


\bibitem[\protect\citeauthoryear{{Wang}, {Gjoreski}, {Ciliberto}, {Mekki},
  {Valentin}, and {Roggen}}{{Wang} et~al\mbox{.}}{2019b}]%
        {RoggenDatasetAndSurvey}
\bibfield{author}{\bibinfo{person}{L. {Wang}}, \bibinfo{person}{H. {Gjoreski}},
  \bibinfo{person}{M. {Ciliberto}}, \bibinfo{person}{S. {Mekki}},
  \bibinfo{person}{S. {Valentin}}, {and} \bibinfo{person}{D. {Roggen}}.}
  \bibinfo{year}{2019}\natexlab{b}.
\newblock \showarticletitle{Enabling Reproducible Research in Sensor-Based
  Transportation Mode Recognition With the Sussex-Huawei Dataset}.
\newblock \bibinfo{journal}{\emph{IEEE Access}}  \bibinfo{volume}{7}
  (\bibinfo{year}{2019}), \bibinfo{pages}{10870--10891}.
\newblock
\showISSN{2169-3536}
\urldef\tempurl%
\url{https://doi.org/10.1109/ACCESS.2019.2890793}
\showDOI{\tempurl}


\bibitem[\protect\citeauthoryear{Zebin, Sperrin, Peek, and Casson}{Zebin
  et~al\mbox{.}}{2018}]%
        {lstmApproach}
\bibfield{author}{\bibinfo{person}{Tahmina Zebin}, \bibinfo{person}{Matthew
  Sperrin}, \bibinfo{person}{Niels Peek}, {and} \bibinfo{person}{Alex Casson}.}
  \bibinfo{year}{2018}\natexlab{}.
\newblock \showarticletitle{Human activity recognition from inertial sensor
  time-series using batch normalized deep LSTM recurrent networks}. In
  \bibinfo{booktitle}{\emph{IEEE EMBC}}.
\newblock
\urldef\tempurl%
\url{https://doi.org/10.1109/embc.2018.8513115}
\showDOI{\tempurl}


\bibitem[\protect\citeauthoryear{Zheng, Fu, Xie, Ma, and Li}{Zheng
  et~al\mbox{.}}{2011}]%
        {zheng2011geolife}
\bibfield{author}{\bibinfo{person}{Yu Zheng}, \bibinfo{person}{Hao Fu},
  \bibinfo{person}{Xing Xie}, \bibinfo{person}{Wei-Ying Ma}, {and}
  \bibinfo{person}{Quannan Li}.} \bibinfo{year}{2011}\natexlab{}.
\newblock \bibinfo{booktitle}{\emph{Geolife GPS trajectory dataset - User
  Guide} (\bibinfo{edition}{geolife gps trajectories 1.1} ed.)}.
\newblock
\urldef\tempurl%
\url{https://www.microsoft.com/en-us/research/publication/geolife-gps-trajectory-dataset-user-guide/}
\showURL{%
\tempurl}
\newblock
\shownote{Geolife GPS trajectories 1.1.}


\end{thebibliography}

%
%
%
%
%
%
%
%

\end{document}